\documentclass[letterpaper, 10 pt, conference]{ieeeconf} 
\IEEEoverridecommandlockouts  
\usepackage{graphicx} % Required for inserting images
\usepackage{soul,color}
\usepackage{amsmath}
\usepackage{amsfonts}

\title{Active Anomaly Detection in Confined Spaces Using \\ Ergodic Traversal of Directed Region Graphs}

\author{Benjamin Wong$^{1}$, Tyler M. Paine$^{2}$, Santosh Devasia$^{1}$, and Ashis G. Banerjee$^{1,3}$
\thanks{This work is supported by a Naval Engineering Education Consortium (NEEC) award \# N00174-22-1-0013.}
\thanks{$^{1}$B. Wong and S. Devasia are with the Department of Mechanical Engineering, University of Washington, Seattle, WA 98195, USA,
        {\tt\small bycw,devasia@uw.edu}}%
\thanks{$^{2}$T. M. Paine is with the Naval Undersea Warfare Center, %Division Keyport, 
Keyport, WA 98345, USA,
        {\tt\small tyler.m.paine.civ@us.navy.mil}}%
\thanks{$^{3}$A. G. Banerjee is with the Department of Industrial \& Systems Engineering and the Department of Mechanical Engineering, University of Washington, Seattle, WA 98195, USA,
        {\tt\small ashisb@uw.edu}}%
}\date{July 2023}

\begin{document}

\maketitle

\begin{abstract}
We provide the first step toward developing a hierarchical control-estimation framework to actively plan robot trajectories for anomaly detection in confined spaces. The space is represented globally using a directed region graph, where a region is a landmark that needs to be visited (inspected). We devise a fast mixing Markov chain to find an ergodic route that traverses this graph so that the region visitation frequency is proportional to its anomaly detection uncertainty, while satisfying the edge directionality (region transition) constraint(s). Preliminary simulation results show fast convergence to the ergodic solution and confident estimation of the presence of anomalies in the inspected regions. 
\end{abstract}

\section{Introduction}
In our previous work, we developed a probabilistic mapping method for human-assisted detection of foreign objects debris in confined spaces toward the goal of inspecting large marine vessels \cite{WongBenjamin2023Hrdo}. Here, we extend this work for more general-purpose anomaly detection (e.g., corroded components, damaged pipes, etc.) that requires the robot to plan trajectories actively. 
To this end, we adopt the idea of ergodic exploration \cite{MavrommatiAnastasia2018RACa} to ensure adequate coverage of the entire space. However, existing ergodic exploration techniques yield 
complex trajectories that are hard to execute in narrow spaces, and introduce non-linearities in the simultaneous localization and mapping (SLAM) pose graph optimization problem. We instead investigate the feasibility of a dual control-estimation hierarchical framework to achieve globally ergodic coverage while maintaining smooth and easily realizable local trajectories.

%\section{Dual-Hierarchical Framework}
%\begin{figure}[thpb]
      %\centering
      %\framebox{\parbox{0.4\textwidth}{ \includegraphics[width=0.4\textwidth]{Figures/framework.pdf}}
%}
      %\caption{A hierarchical framework for ergodic inspection}
      %\label{flowchart}
   %\end{figure}
   
%Our dual hierarchical framework is shown in Figure \ref{flowchart}. The estimation module is inspired by Kimera \cite{RosinolAntoni2019KaOL} and Atlas \cite{BosseM.2003AAff}, and is adapted for inspection purposes. The planner-controller module, being a dual of the estimation module, inherits the same advantages. 
In this paper, we focus on the development of an ergodic Markov chain-based planner that sits at the top of the hierarchy. The states of this chain correspond to confined space regions, defined based on the visibility of the structures that need to be inspected. This defines an underlying directed graph with the edges connecting the regions (nodes) that allow robot transitions. We then define our own ergodic metric such that the frequency of visiting (inspecting) each region is proportional to the anomaly detection uncertainty of that region. The inspection or traversal sequence is obtained by solving for a Markov transition matrix that minimizes the mixing time. Preliminary simulation results show that the computed sequences enable a ground robot to be more confident about the presence of anomalies in the visited regions than random traversal or maximum entropy based exploration. 

%The waypoint planner generates an initial trajectory by maximizing the information gain using Monte Carlo tree search, and navigates to the next (connected) region using RRT. This trajectory is then optimized using sequential quadratic programming based on a second order approximation of the cost map. Last, waypoint navigation is performed using a combination of spline-based feedforward and LQR-based feedback tracking.

% red is Prof. Banerjee or Santosh -- dont remove it and if addressed put a note next to it
% cyan: comments from Prof. Banerjee or Santosh  --- remove color after you address it.
% blue: changes by Ben that needs to be reviewed and blue color removed by either Prof. Banerjee or Santosh  

\section{Anomaly Detector}
We extend our previous anomaly detection algorithm to generate a metric to guide the ergodic exploration. The anomaly detection is applied once the pose graph is available from any SLAM solution. For each point in the point cloud associated with a pose node, the uncertainty of the point is propagated from the localization uncertainty and mapping uncertainty as 
\begin{equation}
\Sigma_p=\frac{\partial h}{\partial X}\Sigma_X\frac{\partial h}{\partial X}^T+R \Sigma_{p/X} R^T .
\end{equation}
Here $X$, \(\Sigma_{X}\) are the localization of the pose node and its corresponding covariance matrix, available from the graph-SLAM. \(\Sigma_{p/X}\) is the covariance matrix of the point with respect to the node, available from the EKF-SLAM; $h$ is the rigid transformation that map the point from the node's reference frame to the global frame. In the 2-D case:
\begin{equation}
h(p)=\begin{bmatrix}
\cos(X_\theta) & -\sin(X_\theta) & X_x\\
\sin(X_\theta) & \cos(X_\theta) & X_y\\
0 & 0 & 1
\end{bmatrix}\begin{bmatrix}
\cos(p_x) \\
\sin(p_y) \\
 1
\end{bmatrix}
\end{equation}

\begin{equation}
\frac{\partial h}{\partial X}=\begin{bmatrix}
1 & 0 & -p_y\cos(X_\theta)-p_x\sin(X_\theta)\\
0 & 1 & p_x\cos(X_\theta)-p_y\sin(X_\theta)\\
\end{bmatrix}
\end{equation}

\begin{equation}
R=\begin{bmatrix}
\cos(X_\theta) & -\sin(X_\theta)\\
\sin(X_\theta) & \cos(X_\theta)\\
\end{bmatrix}
\end{equation}

After acquiring a Gaussian estimation for all the points, we use Bayesian hypothesis testing as an anomaly detector. First, we discretize the reference mesh model into a point cloud, inheriting the normal vector \(n\) from the plane they were sampled from. Then for each point in the online map, we find the closest reference point \(\mu_{ref}\). The halfspace generated by the plane intersecting the reference point forms our null hypothesis \(H_0\): the point is sampled from the reference plane or behind. We interpret this as no anomaly. In particular, we calculate the smallest Mahalanobis distance from the point to the halfspace
\begin{equation}
\mu_0=\begin{cases}
\mu_0^* & \text{ if } n\cdot p>n\cdot \mu_{ref}\\ 
p & \text{ otherwise }  .
\end{cases}
\end{equation}
\(\mu^*\) is the statistically closest point from the query point to the plane, which can be found by the optimization
%\begin{equation}
%\mu^*=\begin{array}{rl}
%\displaystyle \text{arg} \min_{\mu} & (p-\mu)^T\Sigma_p^{-1}(p-\mu)\\
%\textrm{s.t.} & n\mu=n\mu_{ref}\quad .\\
%\end{array}
%\end{equation}
\begin{equation}
\begin{array}{rl}
\displaystyle \text{arg} \min_{\mu} & (p-\mu)^T\Sigma_p^{-1}(p-\mu)\\
\textrm{s.t.} & n\mu=n\mu_{ref}\quad .\\
\end{array}
\end{equation}
The closed form solution is then found using the Lagrange multiplier as
\begin{equation}
\begin{bmatrix}
-2\Sigma_p^{-1} & n^T\\
n & 0
\end{bmatrix}\begin{bmatrix}
\mu_0^*\\
\lambda
\end{bmatrix}=\begin{bmatrix}
-2\Sigma_p^{-1}p\\
n\cdot \mu_0
\end{bmatrix}
\end{equation}
The smallest squared Mahalanobis distance is calculated by
\begin{equation}
D_{0}^*=(p-\mu_0)^T\Sigma_p^{-1}(p-\mu_0)
\end{equation}
Using all the squared Mahalanobis distances of the $k$ neighborhood points gives us the likelihood
\begin{equation}
P(p|H_0)=P\left(\chi_{3k}^2>\sum_i^k D_{0,i}\right)
\end{equation}

We define the alternative hypothesis by offsetting the reference plane outward by some threshold value \(\epsilon\). This threshold represents the allowable deviation from the reference model that is not considered anomaly. In other words, the alternative hypothesis is \(H_1\): the point is sampled from the halfspace outside of the plane of allowable deviation. The likelihood \(P(p|H_1) \) is found be repeating the same procedure with the offset plane. 
\begin{equation}
\mu_1=\begin{cases}
\mu_1^* & \text{ if } n\cdot p<n\cdot \mu_{ref}+\epsilon\\ 
p & \text{ otherwise }  .
\end{cases}
\end{equation}
\begin{equation}
\begin{bmatrix}
-2\Sigma_p^{-1} & n^T\\
n & 0
\end{bmatrix}\begin{bmatrix}
\mu_1^*\\
\lambda
\end{bmatrix}=\begin{bmatrix}
-2\Sigma_p^{-1}p\\
n\cdot \mu_0+\epsilon
\end{bmatrix}
\end{equation}
\begin{equation}
D_{1}^*=(p-\mu_1)^T\Sigma_p^{-1}(p-\mu_1)
\end{equation}
\begin{equation}
P(p|H_1)=P\left(\chi_{3k}^2>\sum_i^k D_{1,i}\right)
\end{equation}
\begin{figure}[thpb]
      \centering
      \framebox{\parbox{0.4\textwidth}{ \includegraphics[width=0.4\textwidth]{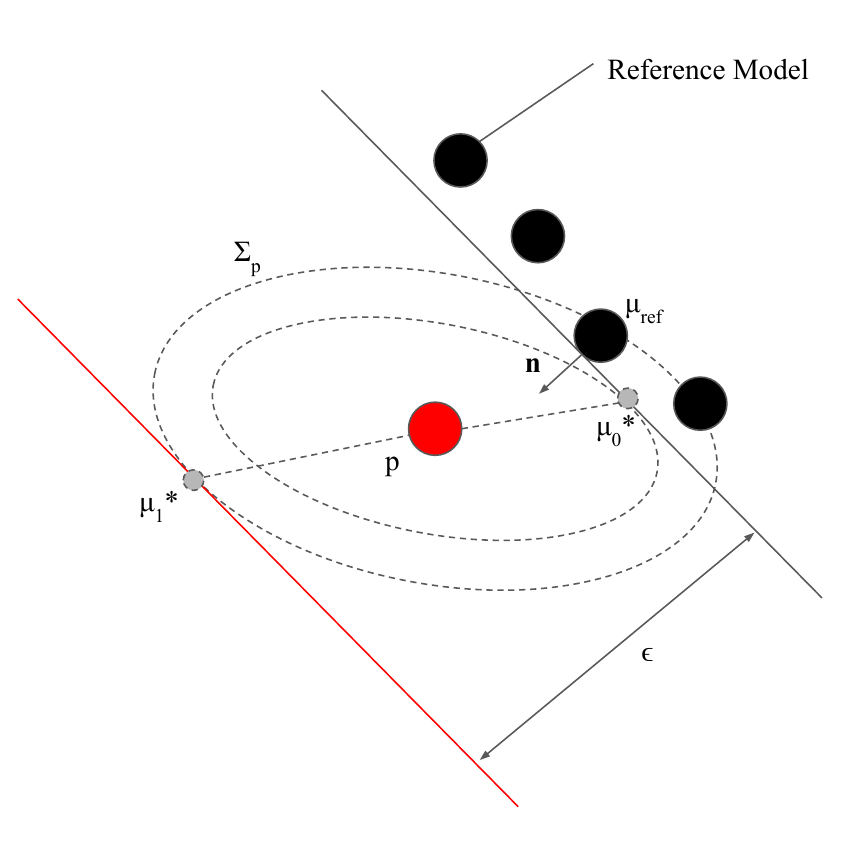}}
}
      \caption{Example sketch illustrating the Bayesian hypothesis testing for a single point \(p\), where the red point is the query point with observation uncertainty \(\Sigma_p\), black points are the point cloud of the reference model, with reference plane specified by the closest point \(\mu_{ref}\) and its normal vector \(n\). \(\mu_0^*\) is the point on the reference plane with maximum log-likelihood. Red plane is the alternative hypothesis plane, generated by offsetting the reference plane in the \(n\) direction by the distance threshold \(\epsilon\), with the corresponding maximum point \(\mu_1^*\).}
      \label{bayesian}
   \end{figure}
We can perform Bayesian update on the competing hypotheses recursively using the likelihood. To prevent the Bayesian belief from collapsing to 0 from extreme observations, we add a smoothing constant $c$ to both the likelihoods.
\begin{equation}
P(H_0| p_{0:t+1})=\eta (P(p_{t+1}|H_0)+c)P(H_0|p_{0:t})
\end{equation}
\begin{equation}
P(H_1|p_{0:t+1})=\eta (P(p_{t+1}|H_1)+c)P(H_1|p_{0:t})
\end{equation}

%\section{Human as Information Source}
%\textcolor{blue}{Every time the robot is near the entrance, it can query the user to identity the anomaly candidates collected during the previous loop. Instead of taking the user input as a definitive label, we instead model the error rate of human labeling as a conditional probability distribution. With the Bayesian hypothesis framework, the robot can incorporate the labeling as an observation using the update rule: 
%\begin{equation}
%P(H_0| z_{\text{h,0:t+1}}, p_{0:t})=\eta P(z_{\text{h,t}}|H_0)\times
%P(H_0| z_{\text{h,0:t}}, p_{0:t})
%\end{equation}
%\begin{equation}
%P(H_1| z_{\text{h,0:t+1}}, p_{0:t})=\eta P(z_{\text{h,t}}|H_1)\times
%P(H_1| z_{\text{h,0:t}}, p_{0:t})
%\end{equation}
%where \(\eta\) is again the normalizing constant such that \(P(H_0)\) and \(P(H_1)\) add up to \(1\); \(z_{\text{h}}\) is a binary label from the human indicating whether a candidate is an anomaly.
%}

\section{Region Sequencer}
%introdction
The first level of the hierarchical planner is the region sequencer. The main objective of the region sequencer is to plan a sequence of regions for the robot to visit.
%in order. 
Moreover, to provide a good balance between information maximation and full coverage, we have adopted the idea of ergodic exploration similar to \cite{MavrommatiAnastasia2018RACa}, where the frequency of visiting each region is proportional to the expected information gain.
If the region sequencer were to focus on information maximization, 
%where the trajectory is found by maximizing the information gain, which can cause
then the robot would tend to over-commit on  the region with the highest information gain and not explore the rest, regardless of how small the difference are. On the other hand with a focus on  full coverage of the graph with minimum repeated nodes, 
the robot would needlessly travel to regions with high prior information.
Moreover, the sequence will be long before the robot revisits the same region, which can have a detrimental effect on the SLAM algorithm of the robot because of the reduced number of loop closures. We propose to solve the ergodic sequencing problem using Markov Chain Monte Carlo (MCMC). Instead of using the popular Metropolis-Hasting algorithm, we have modified the fastest mixing Markov Chain optimization from \cite{BoydStephen2004FMMC} to generate a fast mixing Markov Chain that can converge to any distribution using the upper bound property of L2 norm. This section describes the details of the method.

%region definition 
 \subsection{Region-based directed graph}
For our application, a structure \(\mathcal{S}\) is a subset of points \(m\) with normal vectors from the reference model \(\mathcal{S}:= \{(\mu , n)\quad  | \quad \mu,n \in \mathbb{R}^3,\}\). It represents a collection of points in the workspace that forms a recognizable structure such as a pipe, a wall, an I-beam etc. In this work, the subsets is predefined by user from the CAD model, but it can be done algorithmically with appropriate tools such as instance segmentation. A region \(\mathcal{R}_n\) is defined as a set of robot configurations such that at least one point in the structure \(\mathcal{S}_n\) in the workspace is visible to the robot and the configurations are in the free space. These regions are organized as a graph \(\mathcal{G}\), with the regions as nodes. The edge set \(\mathcal{E}\) is defined such that each edge between two nodes represents there are shared robot 
configurations between the two regions.
These edges can be directed, e.g., due to steep slopes that a ground robot can only traverse in one direction; robot is not allowed to drive backward due to collision risk.
The idea is that once the robot enters the region, it can at least observe part of the structure; if the two regions are connected on the graph, the robot can seamlessly transfer from observing one structure to another. 

%region definition 
\subsection{Goal: Ergodic region sequencer}
The goal for the region sequencer is to traverse the region graph in a way such that the frequency of visiting each node is proportional to the expected information gain of the the region. The assumption is that upon arriving, the robot collect observations, and the belief of the robot converges toward the truth regardless of the prior. This forms our ergodic objective
\begin{equation}
\min_{\mathcal{X}}\quad \left\|\label{ergodic_metric}
\frac{1}{K}\sum_{k=0}^{K-1} \delta(\mathcal{X}[k]-\mathcal{R}_i)-w_i \right\|_{tv}
\end{equation}
\begin{equation}
\label{}
w_i=\frac{h_i}{\sum_j h_j} ,
\end{equation}
where \(K\) is the length of the sequence; \(\delta\) is the Kronecker delta;  \(\mathcal{X}\) is the sequence of regions; \(w\) is the normalized expected information gain. In particular, we choose the entropy of the hypothesis testing \(h_i\) from the anomaly detector as our information gain metric for each region $\mathcal{R}_i$
\begin{equation}
\label{node_entropy}
h_i=\max_{m \in \mathcal{S}_i} \left[-P_{m}(H_0)\ln(P_{m}(H_0))-P_m(H_1)\ln(P_{m}(H_1))\right]
\end{equation} The goal is to find \(\mathcal{X}\) such that the total variation distance between the relative frequency of the region in the sequence and the uncertainty measure is minimized. 

\subsection{Proposed approach using Markov transition matrix}
We propose to generate the sequence using an ergodic Markov chain. For each directed edge \(e_{ij} \in \mathcal{E} \), we assign a transition probability \(p_{ij}\). These transition probability can be arranged into the Markov transition matrix \(P\) of the graph. Then the probability distribution of the robot in each node at time step \(k\) can be written using the left hand Markov transition matrix
\begin{equation}
\begin{bmatrix}
p(\mathcal{X}[k]=\mathcal{R}_1)\\
p(\mathcal{X}[k]=\mathcal{R}_2)\\
\vdots \\
p(\mathcal{X}[0]=\mathcal{R}_n)
\end{bmatrix}=P^k\begin{bmatrix}
p(\mathcal{X}[0]=\mathcal{R}_1)\\
p(\mathcal{X}[0]=\mathcal{R}_2)\\
\vdots \\
p(\mathcal{X}[0]=\mathcal{R}_n)
\end{bmatrix}.
\end{equation}
\noindent
The expected value of the Kronecker delta at region $\mathcal{R}_i$, in time $k$, given initial region $\mathcal{R}_j$ is then 
\begin{equation}
\label{,}
\mathbb{E}_{\mathcal{X}[0]=\mathcal{R}_j}\left [ \delta(\mathcal{X}[k]-\mathcal{R}_i) \right ]=P^k_{i,j} .
\end{equation}
For any irreducible and aperiodic Markov chain, the asymptotic behavior follows \cite{meyn_tweedie_glynn_2009}
\begin{equation}
\lim_{k\rightarrow\infty} P^k =  \begin{bmatrix}
\vrule & \vrule & &\vrule \\ 
\pi & \pi & \cdots & \pi \\ 
\vrule  & \vrule & & \vrule
\end{bmatrix},
\end{equation}
where the distribution converges to \(\pi\) regardless of the initial condition. Putting all together, the asymptotic behaviour of the sequence is 
\begin{equation}
\label{stable_distribution}
\lim_{k\rightarrow\infty} \frac{1}{K}\sum_{k=0}^{K-1} \mathbb{E} \left[\delta(\mathcal{X}[k]-\mathcal{R}_i) \right] = \pi_i .
\end{equation}
Comparing Equations (\ref{ergodic_metric}) and (\ref{stable_distribution}), we can observe that the ergodic metric can be minimized asymptotically by designing the matrix $P$ such that 
\begin{equation}
\pi_i=w_i .
\end{equation}

\noindent 
 To generate the sequence, we perform Monte Carlo tree search according to the transition probability until the depth reaches the planning horizon $K$. The sequence with the least TV-distance to the desired distribution is chosen. 

\subsection{Improving convergence to ergodicity}
Since the robot needs to execute the sequence in a finite time window, we also need to consider the convergence rate such that ergodicity is reached as quickly as possible.
The convergence rate of a Markov process is governed by the second largest second largest eigenvalue modulus (SLEM).
The exact solution of fastest mixing problem for a uniform distribution is solved by Boyd \textit{et al.} \cite{BoydStephen2004FMMC}. In our case, the target distribution is not necessarily uniform, which means that the transition matrix is not necessarily doubly stochastic. Hence, we have modified the formulation to approximately solve the non-uniform fastest mixing problem by minimizing the upper bound of the SLEM.

An approach to maximize convergence is to minimize the SLEM after removing the largest eigenvalue.  For example,  the Wielandt deflation  sets the dominant eigenvalue to 0 (even for non-symmetric matrices) \cite{cdi_unpaywall_primary_W1879119993}, 
since the eigenvalues $\lambda$  of  matrix $A-\alpha \hat{v}_1u_1^T $ are 
\begin{equation}
\lambda(A-\alpha \hat{v}_1u_1^T ) = \{\lambda_1-\alpha, \lambda_2 , \lambda_3 \cdots \},
\label{eigenvalues_deflation}
\end{equation}
where $\alpha$ is a scalar, \(\hat{v}_1\) is the eigenvector of matrix \(A\) and \(u_1\) is any vector such that 
\begin{equation}
u_1^T\hat{v}_1=1 .
\end{equation}
This relation is automatically satisfied if \(u_1^T\) is chosen to be the associated left eigenvector, moreover, the deflated matrix will preserve all the eigenvectors of the original matrix.

For any Markov matrix, the largest eigenvalue is  $\lambda_1 = 1 $ and the associated right eigenvector \(\hat{v}_1=w\). The associated left eigenvector \(u_1^T=\mathbf{1}^T\) due to the fact that all the columns must sum to 1. 
Then, selecting $\alpha = \lambda_1 = 1$,  the  matrix deflation is achieved with 
\begin{equation}
\tilde{P}=P-w \mathbf{1}^T,
\end{equation}
where the largest eigenvalue modulus of \(\tilde{P}\) is the SLEM of \(P\).

\vspace{0.1in} 
\subsection{Minimizing the maximum singular value}
Minimizing the maximum eigenvalue modulus of the deflated matrix  is still challenging since it is not a convex problem: 
while the largest eigenvalue of a symmetric matrix is convex, it is not necessarily convex for a general asymmetric matrix. One solution is to  approximately minimize the  largest eigenvalue modulus of \(\tilde{P}\) by utilizing the bounding property of singular values
\begin{equation}
\min_i \sigma_i \leq \min_i |\lambda_i| \leq \max_i  |\lambda_i| \leq \max_i \sigma_i .
\end{equation}

Since the largest singular value is also the 2-Norm of a matrix, this allow us to formulate the convex optimization problem as (referred to as the upper-bound method)
\begin{equation}
\begin{array}{rlll}
\displaystyle \text{arg} \min_{p_{i,j}} & \| \tilde{P} \|_2\\
\textrm{s.t.}   & \tilde{P}=P-w\mathbf{1}^T \\
                & Pw=w \\
                &\mathbf{1}^TP=\mathbf{1}^T\\
                & p_{i,j} \geq 0   & \forall i,j\\
                & p_{i,j} = 0 & \forall (i,j) \notin \mathcal{E}.
\end{array} 
\label{Weilandt_algorithm} 
\end{equation}
Here, the first constraint is the deflation equation; second constraint is the condition for the Markov chain to converge to the target distribution; third constraint is the property of a Markov matrix that all the columns sum to 1; fourth constraint is the property that all the transition probabilities must be non negative; and the last constraint is the fact that transition cannot happen between nodes that are not connected. Essentially, we have removed the symmetric constraint from Boyd's method, extended to any target distribution \(w\) using Wielandt deflation and formulated an approximate optimization using upper-bound property of singular value.

\vspace{0.1in} 
\subsection{Exact minimization of SLEM} 
In some special cases, exact minimization of the SLEM for non-uniform target distribution can be done by  minimizing a singular value problem. For example, if \(P\) is specified to be a transition matrix of a reversible Markov chain, i.e., which satisfies the detailed-balance 
constraint
\begin{equation}
\begin{aligned} 
P W & = W P^T 
\end{aligned} 
\label{equation_reversibility}
\end{equation}
with $W  =\text{diag}(w)$, exact minimization of SLEM is possible using  Boyd's algorithm~\cite{BoydStephen2004FMMC} 
(referred to as the  fastest mixing reversible Markov chain (FMRMC) method)
\begin{equation}
\begin{array}{rlll}
\displaystyle \text{arg} \min_{p_{i,j}} & \| \tilde{P} \|_2\\
\textrm{s.t.}   & \tilde{P}=W^{-1/2}PW^{1/2}-qq^T \\
                & PW=W P^T \\
                &\mathbf{1}^TP=\mathbf{1}^T\\
                & p_{i,j} \geq 0   & \forall i,j\\
                & p_{i,j} = 0 & \forall (i,j) \notin \mathcal{E},
\end{array} 
\label{Boyds_algorithm} 
\end{equation}
where  \(q=\begin{bmatrix}
\sqrt{w_1} & \sqrt{w_2} & \cdots  & \sqrt{w_n}
\end{bmatrix}^T\).

\noindent 
%When the target distribution is uniform $w =\mathbf{1}/n $, 
%then the two algorithms (in Eqs.~\eqref{Weilandt_algorithm} and \eqref{Boyds_algorithm}) become the same for undirected graphs, Eq.~\eqref{equation_reversibility}. 

 \subsection{Proposed algorithm }
\noindent 
In general, the reversibility condition 
 \begin{equation}
 P_{i,j} w_j = P_{j,i}w_i  \end{equation} 
 is more conservative than the stable distribution condition \(Pw=w\) as it set additional constraints to each connected node pairs. This is more apparent when applied to a directed graph, as it sets both the edge weights to zero 
  \begin{equation}
  P_{i,j} = P_{j,i} =0,
  \label{eq_edge_removal}
   \end{equation} 
   if either one $  P_{i,j}$ or $ P_{j,i}$ needs to be zero because the edge is directed. 
However, we empirically observed that for undirected graph, reversibility leads to lower SLEM than it's irreversible counterparts. From this observation, we propose to apply the same 
similarity transformation from section 6 of \cite{BoydStephen2004FMMC}. while avoid the reversibility condition. The objective turns into a upper-bound minimization instead of the exact eigenvalue minimization similar to the previous upper-bound method. 
i.e., 
(referred to as the modified upper-bound method)
 \begin{equation}
\begin{array}{rlll}
\displaystyle \text{arg} \min_{p_{i,j}} & \| \tilde{P} \|_2\\
\textrm{s.t.}   & \tilde{P}=W^{-1/2}PW^{1/2}-qq^T \\
                & Pw=w \\
                &\mathbf{1}^TP=\mathbf{1}^T\\
                & p_{i,j} \geq 0   & \forall i,j\\
                & p_{i,j} = 0 & \forall (i,j) \notin \mathcal{E}.
\end{array} 
\label{Ben_modified_Boyd}
\end{equation}

We observe via simulation that the modified algorithm in ~\eqref{Ben_modified_Boyd} also finds the fastest mixing reversible Markov chain as in \eqref{Boyds_algorithm}, but without applying the detailed-balance 
constraint in Eq.~\eqref{equation_reversibility}. Moreover, for directed graphs, the modified algorithm in ~\eqref{Ben_modified_Boyd} outperforms the fastest mixing reversible Markov chain algorithm in~\eqref{Boyds_algorithm}, since the reversibility condition effectively leads to the removal of the directed edges according to Eq.~\eqref{eq_edge_removal}.  %{\color{red}Ben: Please add the Monte Carlo tree search-based planning part that you had earlier.}

\section{Preliminary Results}
\begin{figure}[thpb]
      \centering
      \framebox{\parbox{0.4\textwidth}{ \includegraphics[width=0.4\textwidth]{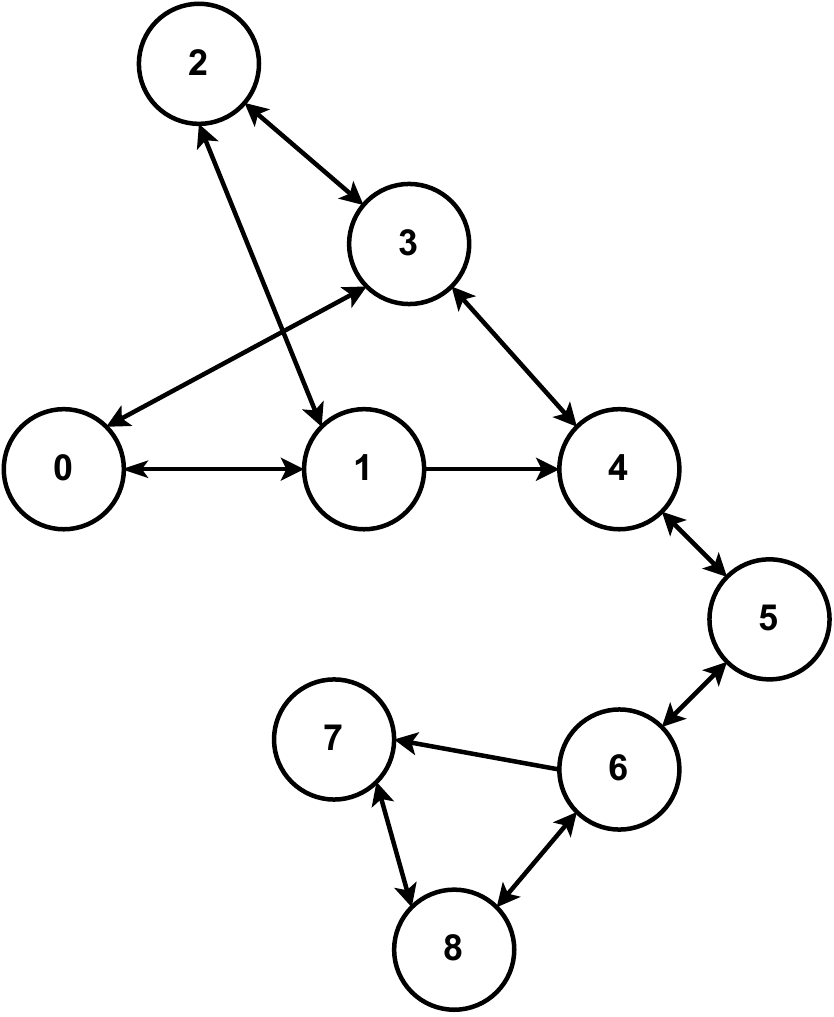}}
}
      \caption{An example graph representation of 9 regions, with edges 1-4 and 6-7 being directed.} 
      \label{example_graph}
   \end{figure}
   
\subsection{Fast mixing Markov Chain}
To compare the various ergodic Markov chain algorithms, we construct an example graph, inspired by a confined workspace (water tank) in a marine vessel, as shown in Figure \ref{example_graph}. Each entry \(w_i\) of the target distribution \(w\) is randomly sampled from the range \([0, 1]\), and normalized such that the sum of the entries is \(1\) for a total of 1000 trials. We first test the results on a fully undirected version of the example graph. 

\begin{figure}[thpb]
      \centering
      \framebox{\parbox{0.45\textwidth}{ \includegraphics[width=0.45\textwidth]{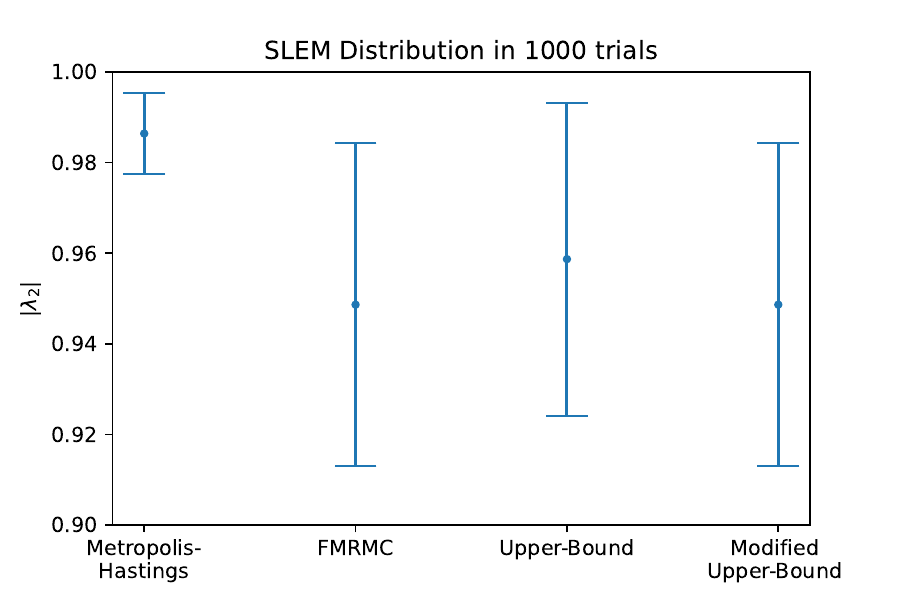}}
}
      \caption{Distribution of SLEM for each algorithm in 1000 trials with randomly generated target distributions on the undirected version of the graph in Fig.~\ref{example_graph}. Modified upper-bound method matches the FMRMC method for undirected graphs. 
     }
      \label{undirectedSLEM}
   \end{figure}

Figure \ref{undirectedSLEM} shows the distribution of the SLEM of the transition matrices generated by Metropolis-Hastings algorithm (Section 1.2.2. of \cite{BoydStephen2004FMMC}), FMRMC (Eq. \ref{Boyds_algorithm}), upper-bound method (Eq. \ref{Weilandt_algorithm}), and modified upper-bound method (Eq. \ref{Ben_modified_Boyd}). The result shows that Metropolis-Hastings algorithm generates the worst SLEM on an average. Despite the addition of redundant %unnecessary 
detailed-balance constraints, FMRMC still outperforms our upper-bound method. However, our modified upper-bound method, which uses the same similarity transform as the FMRMC, yields identical SLEM for all the trials. 

%\begin{figure}[thpb]
 %     \centering
  %    \framebox{\parbox{0.45\textwidth}{ \includegraphics[width=0.45\textwidth]{Figures/final.pdf}}
%}
 %     \caption{Comparison of frequency in 20 time steps between optimization method vs. Metropolis-Hasting method on the example graph with a example target distribution.}
  %    \label{flowchart}
   %\end{figure}
   
%\begin{figure}[thpb]
 %     \centering
  %    \framebox{\parbox{0.45\textwidth}{ \includegraphics[width=0.45\textwidth]{Figures/tvdist.pdf}}
%}
 %     \caption{}
  %    \label{flowchart}
   %\end{figure}

\begin{figure}[thpb]
      \centering
      \framebox{\parbox{0.45\textwidth}{ \includegraphics[width=0.45\textwidth]{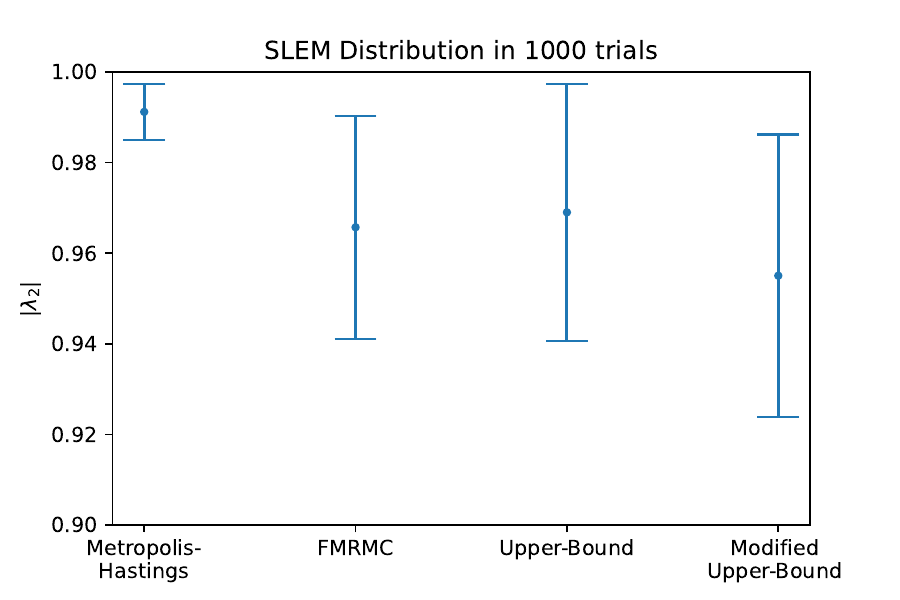}}
}
      \caption{Distribution of SLEM for different traversal methods in 1000 trials with randomly generated target distributions on 
      the directed region graph in Fig.~\ref{example_graph}. 
      Our modified upper-bound method outperforms the well-established FMRMC method. 
      }
      \label{directedSLEM}
   \end{figure}

 We performed the same trials on the directed version of the same graph, with edges 1-4 and 6-7 being directed. The result is shown in Figure \ref{directedSLEM}. All the methods have a decrease in performance with the introduction of directed edges due to a loss in degrees of freedom. This is particularly true for the methods that require detailed-balance, which forces the directed edges to be zero. Since the detailed-balance condition is removed from our upper-bound method, we are able to achieve better performance. With lower SLEM, the Markov chain reaches stationary distribution faster on an average, which, in turn, lowers the expected ergodicity cost of the sequence generated by MCMC in a finite time horizon.

\subsection{Ergodic Exploration on Graph}
\begin{figure}[thpb]
      \centering
      \framebox{\parbox{0.4\textwidth}{ \includegraphics[width=0.4\textwidth]{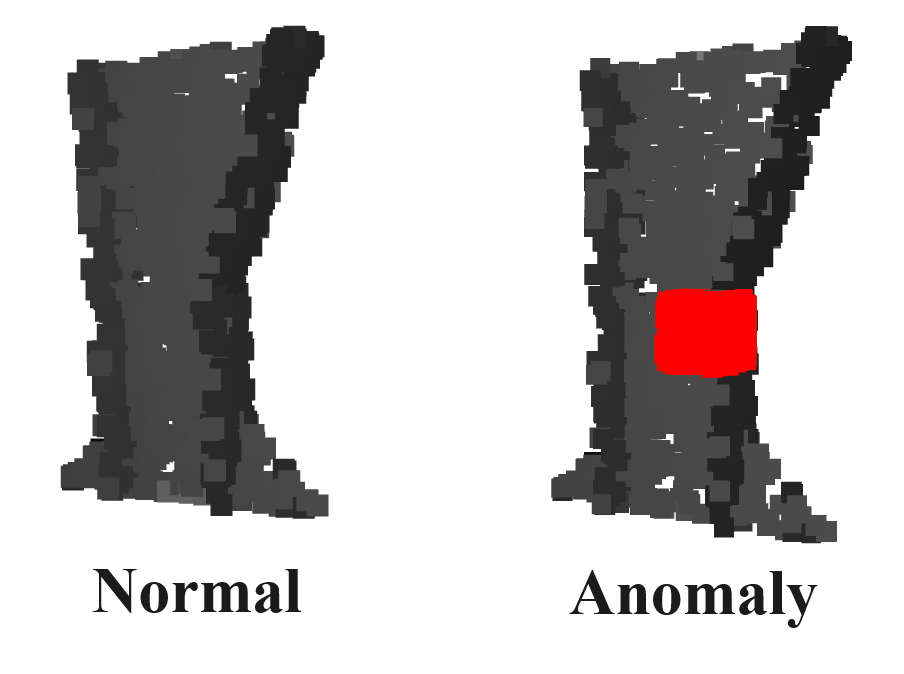}}
}
      \caption{Two types of structure model contained in each node. Left hand side being the default I-beam structure. Right hand side being the anomalous version, with a red cube attached to the default I-beam.}
 
      \label{model}
   \end{figure}

Using the same graph as in Figure \ref{example_graph}, we simulate an anomaly detection scenario to test the ability of finding anomalies for various graph traversal methods. Each node in the graph contains a mesh of an I-beam structure, shown in the left hand side of Figure \ref{model}, with a total number of 500 points. Each node has a chance of containing an anomalous cube of size 100 $\times$ 100 $\times$ 100 units, with 250 points. At the beginning of each trial, the anomaly assignment probability is generated uniformly from 0.25 to 0.75 for each node. The anomalies are then assigned to each node randomly according to this probability. The robot starts at node 0, and samples 250 points every time it arrives at a node with the samples corrupted by zero mean Gaussian noise with \(\Sigma=\text{diag} \left(\begin{bmatrix}
40 & 40 & 40\\ 
\end{bmatrix}\right)\). The samples are immediately processed by the anomaly detector, with threshold \(\epsilon=20 \), smoothing constant \(c=0.5\), and neighborhood number \(k=5\). The robot has a maximum of 20 steps available for traversing the graph. The prior beliefs are generated by perturbing the anomaly assignment probability with a Gaussian noise of \(0.2\). If the value falls outside the range \([0,1]\), it is re-sampled until it comes within the range. This prior is assigned to all the points in the reference structure.

We compared three methods: random, greedy maximum entropy, and entropy ergodic. For the random method, the robot simply moves between neighboring nodes with uniform probability. For the greedy maximum entropy, the robot selects the neighbor with the highest entropy, where the entropy of each node is defined by the reference point with the highest entropy in the structure according to Equation (\ref{node_entropy}). Our ergodic method uses the same entropy measure with a planning horizon \(K=10\). 

\begin{table}[thpb]
\centering
\caption{Average binary cross-entropy losses for the anomalous nodes over 500 trials.}
\label{BCE_anomaly}
\begin{tabular}{|c|c|}
\hline
Method                & BCE loss (anomalous nodes)\\ \hline
Random          & 0.396 $\pm$ 0.01             \\ 
%\hline
Maximum Entropy & 0.569 $\pm$ 0.01           \\ 
%\hline
Entropy Ergodic & \textbf{0.355 $\pm$ 0.01}             \\ 
\hline
\end{tabular}
\end{table}

Table \ref{BCE_anomaly} reports the binary cross entropy (BCE) loss of the \(P(H_1)\) labeling of all the three methods with respect to 1 for the nodes containing actual anomalies at the end of 20 steps over 500 trials. We observe that our ergodic method achieves significantly lower BCE-loss, which means that it is more confident about an anomaly existing in a node based on its traversal sequence. Table \ref{BCE_all} shows the BCE loss of all the nodes with the same setup. The ergodic method again has the lowest loss among the three methods, although the difference in performance is less pronounced than for the anomalous nodes. 

\begin{table}[thpb]
\centering
\caption{Average binary cross-entropy losses for all the nodes over 500 trials.}
\label{BCE_all}
\begin{tabular}{|c|c|}
\hline
Method                & BCE loss (all nodes) \\ \hline
Random          & 0.532 $\pm$ 0.009            \\ 
%\hline
Maximum Entropy & 0.622 $\pm$ 0.009          \\ 
%\hline
Entropy Ergodic & \textbf{0.525 $\pm$ 0.009}             \\ 
\hline
\end{tabular}

\end{table}
\section{Conclusions and Future Work}

In this paper, we provide the first step toward general-purpose anomaly detection in confined spaces by actively planning the motion of an autonomous ground robot. Specifically, we present a new Markov chain Monte Carlo method for discrete, global planning through ergodic traversal of a directed graph that represents the regions that need to be inspected. Preliminary simulation results promising performance in terms of fast convergence to the target (region) visitation distribution and confident estimation of the presence of anomalies in the regions. 

As noted before, this work is expected to form the basis of an active SLAM framework for accurate and robust anomaly detection. In this regard, we plan to come up with a hierarchical, dual estimation-planning framework. The estimation module will include the proposed anomaly detector at the top level, and adapt the Kimera \cite{RosinolAntoni2019KaOL} or Atlas \cite{BosseM.2003AAff} SLAM solutions for the lower level pose graph optimization and robot localization tasks. The anomaly detector can be further enhanced using the state-of-the-art THOR framework \cite{samani2023persistent} for enhanced robustness in unseen environments. For lower level planning, a combination of RRT and sequential quadratic programming can be used for waypoint placement, and spline-based feedforward and LQR-based feedback tracking can be employed for waypoint navigation.

%frameworks, but will be adapted for inspection purposes. The planning module will 
%being a dual of the estimation module, will inherit the same advantages. 

\bibliographystyle{IEEEtran} 
\bibliography{references}

\end{document}